\documentclass[letterpaper]{article} %DO NOT CHANGE THIS
\usepackage{aaai18}  %Required
\usepackage{times}  %Required
\usepackage{helvet}  %Required
\usepackage{courier}  %Required
\usepackage{url}  %Required
\usepackage{graphicx}  %Required
\usepackage{fancyvrb}
\usepackage{csquotes} %added by Larry

\begin{document}
\title{Rationalization: A Neural Machine Translation Approach to Generating Natural Language Explanations}
% The \author macro works with any number of authors. There are two
% commands used to separate the names and addresses of multiple
% authors: \And and \AND.
%
% Using \And between authors leaves it to LaTeX to determine where to
% break the lines. Using \AND forces a line break at that point. So,
% if LaTeX puts 3 of 4 authors names on the first line, and the last
% on the second line, try using \AND instead of \And before the third
% author name.

% \author{Upol Ehsan \thanks{Harrison and Ehsan equally contributed to this work}\\
% Georgia Institute of Technology\\
% Atlanta, GA, USA\\
% \texttt{ehsanu@gatech.edu}\\
% \And
% Brent Harrison \footnotemark[1]\\
% University of Kentucky\\
% Lexington, KY, USA\\
% \texttt{harrison@cs.uky.edu}\\
% \And
% Larry Chan \\
% Georgia Institute of Technology\\
% Atlanta, GA, USA\\
% \texttt{larrychan@gatech.edu}\\
% \And
% Mark Riedl \\
% Georgia Institute of Technology\\
% Atlanta, GA, USA\\
% \texttt{riedl@cc.gatech.edu}\\
% }
\author{
Upol Ehsan \thanks{Harrison and Ehsan equally contributed to this work}$^{\dagger}$, 
Brent Harrison \footnotemark[1]$^\ddagger$, 
Larry Chan$^{\dagger}$, 
{\normalfont and} Mark Riedl$^{\dagger}$\\
$^\dagger$Georgia Institute of Technology, Atlanta, GA, USA\\
$^\ddagger$University of Kentucky, Lexington, KY, USA \\
\texttt{ehsanu@gatech.edu}, 
\texttt{ harrison@cs.uky.edu}, 
\texttt{ larrychan@gatech.edu}, 
\texttt{ riedl@cc.gatech.edu}
}
% \author{XXXX}
  %% examples of more authors
%   \And
  %% Coauthor \\
  %% Affiliation \\
  %% Address \\
  %% \texttt{email} \\
  %% \AND
  %% Coauthor \\
  %% Affiliation \\
  %% Address \\
  %% \texttt{email} \\
  %% \And
  %% Coauthor \\
  %% Affiliation \\
  %% Address \\
  %% \texttt{email} \\
  %% \And
  %% Coauthor \\
  %% Affiliation \\
  %% Address \\
  %% \texttt{email} \
  
\maketitle

\begin{abstract} 
% \todo{1 Paragraph, 4-6 sentences}
We introduce {\em AI rationalization}, an approach for generating explanations of autonomous system behavior as if a human had performed the behavior.
We describe a rationalization technique that uses neural machine translation to translate internal state-action representations of an autonomous agent into natural language.
We evaluate our technique in the Frogger game environment, training an autonomous game playing agent to rationalize its action choices using natural language. 
A natural language training corpus is collected from human players thinking out loud as they play the game.
We motivate the use of rationalization as an approach to explanation generation and show the results of two experiments evaluating the effectiveness of rationalization. 
Results of these evaluations show that neural machine translation is able to accurately generate rationalizations that describe agent behavior, and that rationalizations are more satisfying to humans than other alternative methods of explanation. 

\end{abstract} 

\section{Introduction}

Autonomous systems must make complex sequential decisions in the face of uncertainty.
{\em Explainable AI} refers to artificial intelligence and machine learning techniques that can provide human understandable justification for their behavior.
With the proliferation of AI in everyday use, explainability is important in situations where human operators work alongside autonomous and semi-autonomous systems because it can help build rapport, confidence, and understanding between the agent and its operator.
%These qualities becomes important in human-agent (or robot) teams wherein a human operator works alongside an autonomous or semi-autonomous agent.
For instance, a non-expert human collaborating with a robot for a search and rescue mission requires confidence in the robot's action. 
%Since each move in such a complex task will not necessarily be the most optimal one, explainability of the AI agent's decision-making process can promote meaningful interaction with the robot.
%In many cases a human operator will be teamed with an autonomous or semi-autonomous agent or robot.
In the event of failure---or if the agent performs unexpected behaviors---it is natural for the human operator \textit{to want to know why}.
Explanations help the human operator understand why an agent failed to achieve a goal or the circumstances whereby the behavior of the agent deviated from the expectations of the human operator.
They may then take appropriate remedial action: trying again, providing more training to machine learning algorithms controlling the agent, reporting bugs to the manufacturer, etc. 

Explanation differs from {\em interpretability}, which is a feature of an algorithm or representation that affords inspection for the purposes of understanding behavior or results. 
%For example, decision trees are considered to be relatively interpretable and neural networks are generally considered to be uninterpretable without additional processes to visualize patterns of neuron activation \cite{zeiler2014visualizing,yosinskiunderstanding}.
While there has been work done recently on the interpretability of neural networks~\cite{yosinskiunderstanding,zeiler2014visualizing}, these studies mainly focus on interpretability for experts on non-sequential problems. 
Explanation, on the other hand, focuses on sequential problems, is grounded in natural language communication, and is theorized to be more useful for non-AI-experts who need to operate autonomous or semi-autonomous systems.

% Autonomous systems can fail for a number of reasons including:

% \begin{itemize}
% \item Autonomous systems can be given the wrong objective function or can have learned a sub-optimal policy.
% \item Imperfect sensors can causing agents to believe they are in the wrong state and thus perform the wrong behaviors.
% \item Imperfect effectors can cause agents to fail to achieve the desired effects on the world.
% \end{itemize}

In this paper we introduce a new approach to explainable AI: {\em AI rationalization}.
AI rationalization is a process of producing an explanation for agent behavior {\em as if a human had performed the behavior}.
AI rationalization is based on the observation that there are times when humans may not have full conscious access to reasons for their behavior and consequently may not give explanations that literally reveal how a decision was made. 
In these situations, it is more likely that humans create plausible explanations on the spot when pressed.
However, we accept human-generated rationalizations as providing some lay insight into the mind of the other. 

AI rationalization has a number of potential benefits over other explainability techniques:
(1)~by communicating like humans, rationalizations are naturally accessible and intuitive to humans, especially non-experts 
% without AI or computer science training;
(2)~humanlike communication between autonomous systems and human operators may afford human factors advantages such as higher degrees of satisfaction, confidence, rapport, and willingness to use autonomous systems;
(3)~rationalization is fast, sacrificing absolute accuracy for real-time response, appropriate for real-time human-agent collaboration.
Should deeper, more accurate explanations or interpretations be necessary, rationalizations may need to be supplemented by other explanation, interpretation, or visualization techniques.

% We investigate AI rationalization in the context of reinforcement learning (RL) agents.
% Reinforcement Learning \cite{suttonbarto1998} is a framework for sequential decision-making that is relevant to agents and robots operating in stochastic environments such as the real world. RL has several advantages with respect to interpretability: one can inspect histories of visited states and executed actions, and evaluate the agent's belief about state, transition likelihoods, and the actions considered by the agent at each point in time. For highly complex state spaces, such as those related to observations derived from machine vision systems, Deep Reinforcement Learning (DRL) \cite{mnih2015} uses deep neural networks to approximate the state-action transition probabilities and/or the value of being in states that have never been directly observed. 
% %The reliance on predictions from deep neural networks reduces the interpretability and explainability of reinforcement learning.
% Regardless, explanation requires the decision-making process of a reinforcement learning agent---which may be difficult to understand by non-AI-experts and non computer scientists---to be converted automatically into natural language communication.

We propose a technique for AI rationalization that treats the generation of explanations as a problem of {\em translation} between ad-hoc representations of states and actions in an autonomous system's environment and natural language. 
%Language translation technology using encoder-decoder neural networks (e.g., \cite{luong-pham-manning:2015:EMNLP})  has reached the level of maturity such that it is used in commercial language-to-language translation. 
To do this, we first collect a corpus of natural language utterances from people performing the learning task.
We then use these utterances along with state information to train an encoder-decoder neural network to translate between state-action information and natural language. 

To evaluate this system, we explore how AI rationalization can be applied to an agent that plays the game {\em Frogger}.
%We do this for two reasons.
%First, we aim to show the technique works on an autonomous system without adaptation---we use an arbitrary state representation designed for the game without consideration for rationalization.
This environment is notable because conventional learning algorithms, such as reinforcement learning, do not learn to play Frogger like human players, and our target audience would not be expected to understand the specific information about how an agent learns to play this game or why it makes certain decisions during execution. 
We evaluate our approach by measuring how well it can generate rationalizations that accurately describe the current context of the Frogger environment. 
We also examine how humans view rationalizations by measuring how satisfying rationalizations are compared to other baseline explanation techniques. The contributions of our paper are as follows:
%Most people will probably also not have knowledge on the specific state representation used in these environments.
%Second, we aim to establish a baseline that we can compare to later investigations of AI rationalization applied to state abstractions learned from a deep learning techniques such as a convolutional neural network \textbf{[[this second claim is unclear]]}. \textbf{\em Similar to the search and rescue mission scenario described earlier, the Frogger game environment involves sequential decision-making with sub-optimal moves that may be counter-intuitive from a human-perspective. Insights into behind-the-scenes rationalization of the actions can help establish common ground and meaning-making between the human and AI agent.}

%\begin{figure}[tb]
%\centering
%\includegraphics[scale=1.0]{images/Frogger_gameEnvironment.png}
%\caption{The Frogger game environment.}
%\label{fig:frogger}
%\end{figure}

\begin{itemize}
\item We introduce the concept of {\em AI rationalization} as an approach to explainable AI.
\item We describe a technique for generating rationalizations that treats explanation generation as a language translation problem from internal state to natural language.
\item We report on an experiment using semi-synthetic data to assess the accuracy of the translation technique.
\item We analyze how types of rationalization impact human satisfaction and use these findings to inform design considerations of current and future explainable agents.
\end{itemize}

% This paper introduces the concept of AI rationalization, describes our methodology for AI rationalization, and provides an initial set of experiments to measure the accuracy of our method.
% This initial set of experiments sets the stage for future research addressing state abstractions in deep reinforcement learning, as well as exploring questions of how AI rationalization influences human perceptions of trustworthiness and rapport in increasingly realistic real-time scenarios.

%The remainder of the paper is organized as follows.
%In Sections \ref{sec:related-work} we overview related work on explainable AI. 
%In Section \ref{sec:rationalization}, we describe our AI rationalization technique.
%Section \ref{sec:experiments} describes specific details of how we used rationalization in experiments designed to assess the accuracy of the technique. 
%In Section \ref{sec:future-work} we propose a number of next steps in the investigation of rationalization as an approach to explainable AI.

%%%%%%%%%%%%%%%%%%%%%%%%%%%%%%%%%%%%%%%%%%%%%%%%%%%%%%%%%%%%%%%%%%%%%%%%%%%%%%

\section{Background and Related Work} 
\label{sec:related-work}

%We need a section on \textit{interpretability} and a section on language generation. 
% The work in this paper largely concerns the broad topic of \textit{interpretability} in machine learning. 
For a model to be interpretable it must be possible for humans to explain why it generates certain outputs or behaves in a certain way. 
Inherently, some machine learning techniques produce models that are more interpretable than others. 
For sequential decision making problems, there is often no clear guidance on what makes a good explanation.
For an agent using $Q$-learning \cite{watkins1992}, for example, explanations of decisions could range from ``the action had the highest $Q$ value given this state'' to ``I have explored numerous possible future state-action trajectories from this point and deemed this action to be the most likely to achieve the highest expected reward according to iterative application of the Bellman update equation.''
%Models such as decision trees~\cite{letham2015interpretable}, generalized additive models~\cite{caruana2015intelligible}, and attention-based approaches~\cite{xu2015show} have the benefit of being naturally interpretable. 
%This is because one need only examine the output to understand the model's decision making process (in the case of rule based approaches such as decision trees) or examine some internal features which reveal which areas of the model's feature space have the greatest influence on prediction (in the case of attention and regression-based approaches). 
%While these techniques are effective, relying solely on them for interpretable models greatly inhibits machine learning researchers by limiting the types of models they can use for learning.

% \begin{figure}
% \centering
% % \includegraphics[width=5in]{images/Frogger_workflow.pdf}
% \includegraphics[width=3.5in]{images/Frogger_workflow_reduced.png}
% \caption{The workflow for our proposed neural machine translation approach to rationalization generation.}
% \label{fig:workflow}
% \end{figure}
\begin{figure*}[tb]
\centering
\includegraphics[width=6in]{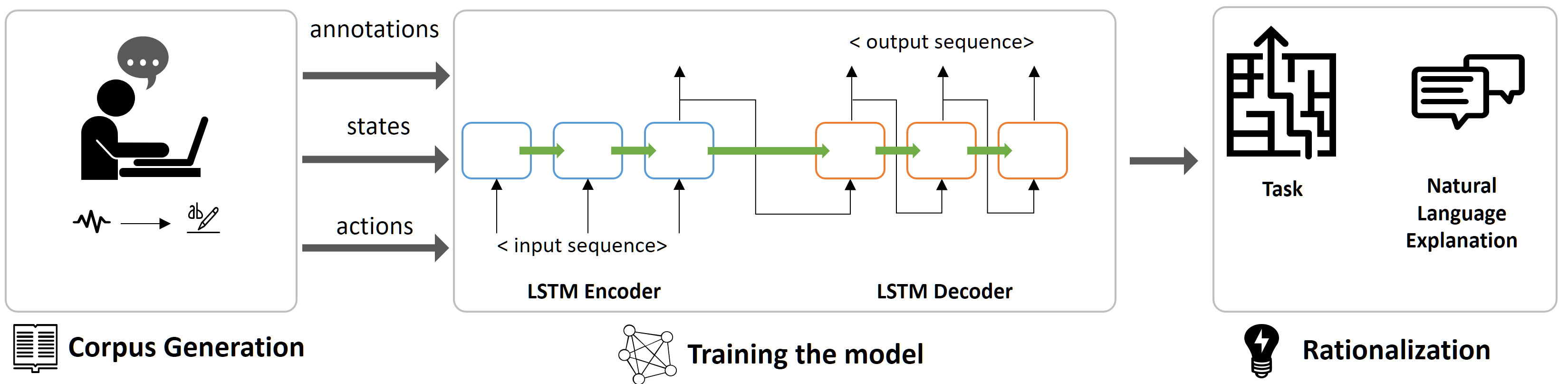}
\caption{The workflow for our proposed neural machine translation approach to rationalization generation.}
\label{fig:workflow}
\end{figure*}

An alternate approach to creating interpretable machine learning models involves creating separate models of explainability that are often built on top of black box techniques such as neural networks. 
These approaches, sometimes called \textit{model-agnostic}~\cite{ribeiro2016should,zeiler2014visualizing,yosinskiunderstanding} approaches, allow greater flexibility in model selection since they enable black-box models to become interpretable.  
Other approaches seek to learn a naturally interpretable model which describes predictions that were made~\cite{krause2016interacting} or by intelligently modifying model inputs so that resulting models can describe how outputs are affected~\cite{ribeiro2016should}.

Explainable AI has been explored in the context of ad-hoc techniques for transforming simulation logs to explanations \cite{vanlent2004}, intelligent tutoring systems \cite{core_building_2006}, transforming AI plans into natural language \cite{vanlent:aiide2005}, and translating multiagent communication policies into natural language \cite{DBLP:journals/corr/AndreasDK17}.
Our work differs in that the generated rationalizations do not need to be truly representative of the algorithm's decision-making process. This is a novel way of applying explainable AI techniques to sequential decision-making in stochastic domains. 
\section{AI Rationalization}
\label{sec:rationalization}
Rationalization is a form of explanation that attempts to justify or explain an action or behavior based on how a human would explain a similar behavior.
Whereas explanation implies an accurate account of the underlying decision-making process, AI rationalization seeks to generate explanations that closely resemble those that a human would most likely give were he or she in full control of an agent or robot.
We hypothesize that rationalizations will be more accessible to humans that lack the significant amount of background knowledge necessary to interpret explanations and that the use of rationalizations will result in a greater sense of trust or satisfaction on the part of the user.
While Rationalizations generated by an autonomous or semi-autonomous system need not accurately reflect the true decision-making process underlying the agent system, they must still give some amount of insight into what the agent is doing. 
Our approach for translating representations of states and actions to natural language consists of two general steps. First, we must create a training corpus of natural language and state-action pairs. Second, we use this corpus to train an encoder-decoder network to translate the state-action information to natural language (workflow in Figure \ref{fig:workflow}).
%We will discuss each of these steps in more detail below. 
%Our system's workflow is shown in Figure .
\subsection{Training Corpus}
Our technique requires a training corpus that consists of state-action pairs annotated with natural language explanations.
To create this corpus, we ask people to complete the agent task's in a virtual environment and ``think aloud'' as they complete the task.
We record the visited states and performed actions along with the natural language utterances of critical states and actions.
%In addition to providing examples of humans completing the task, 
This method of corpus creation ensures that the annotations gathered are associated with specific states and actions. In essence we create parallel corpora, one of which contains state representations and actions, the other containing natural language utterances.
%The state representation and actions constitute a ``machine language''.

%\subsection{Use the network}

The precise representation of states and actions in the autonomous system does not matter as long as they can be converted to strings in a consistent fashion.
%\textbf{Autonomous agent may have a state representation tailored to a specific environment or may use a neural network to learn a representation of the environment.}
Our approach emphasizes that it should not matter how the state representation is structured and the human operator should not need to know how to interpret it.

%The general approach is agnostic to the details of how the parallel corpora are obtained. 
%We describe the specific methodology we used to create the parallel corpora for the purpose of experimentation in Section \ref{sec:methodology}.

\subsection{Translation from Internal Representation to Natural Language}

We use encoder-decoder networks to translate between complex state and action information and natural language rationalizations.
Encoder-decoder networks, which have primarily been used in machine translation and dialogue systems, are a generative architecture comprised of two component networks that learn how to translate an input sequence $X = (x_1, ..., x_T)$ into an output sequence $Y = (y_1, ..., y_{T'})$. 
The first component network, the encoder, is a recurrent neural network (RNN) that learns to encode the input vector $X$ into a fixed length context vector $v$.
This vector is then used as input into the second component network, the decoder, which is a RNN that learns how to iteratively decode this vector into the target output $Y$.
We specifically use an encoder-decoder network with an added attention mechanism~\cite{luong-pham-manning:2015:EMNLP}.
\section{Experiments}
In this work, we test the following two hypotheses:
\begin{enumerate}
\item Encoder-Decoder networks can accurately generate rationalizations that fit the current situational context of the learning environment and
\item Humans will find rationalizations more satisfying than other forms of explainability
\end{enumerate}
To test these hypotheses, we perform two evaluations in an implementation of the popular arcade game, {\em Frogger}.
We chose Frogger as an experimental domain because computer games have been demonstrated to be good stepping stones toward real-world stochastic environments \cite{laird:aimagazine01,mnih2015}
and because Frogger is fast-paced, has a reasonably rich state space, and yet can be learned optimally without too much trouble.

%%%%%%%%%%%%%%%%%%%%%%%%%%%%%%%%%%%%%%%%%%%%%%%%%%%%%%%%%%%%%%%%%%%%%%%%%%%%%

% \section{Rationalization Generation Experiments}
\label{sec:experiments}

%To evaluate the effectiveness of our approach, we perform two experiments.
%The first experiment 
%The objective of the experiments we describe below is to assess the accuracy of the neural machine translation approach to rationalization.
%
\subsection{Rationalization Generation Study Methodology}
% \subsection{Generating Accurate Rationalizations}
%This first experiment is designed to judge whether our neural machine translation approach to generating rationalizations can produce output that accurately describes behaviors in a virtual world.
Evaluating natural language generation is challenging; utterances can be ``correct'' even if they do not exactly match known utterances from a testing dataset. 
To facilitate the assessment of rationalizations generated by our technique, we devised a technique whereby semi-synthetic natural language was paired against state-action representations internal to an autonomous system.
The semi-synthetic language was produced by observing humans ``thinking out loud'' while performing a task and then creating grammar that reproduced and generalized the utterances (described below).
This enables us to use the grammar to evaluate the accuracy of our system since we can compare the rationalizations produced by our system to the most likely rule that would have generated that utterance in the grammar. Similar approaches involving the use of semi-synthetic corpora have been adopted in scenarios, such as text understanding~\cite{weston2015towards}, where ground truth is necessary to evaluate the system.
%identify the grammar rules during autonomous system execution and compare rationalizations generated by the neural machine translation network to a ground-truth. 
%
%\textbf{Later, we can replace the grammar with a corpus entirely trained on humans performing a task in the wild.}

\begin{figure}
\centering
\includegraphics[width= 3in]{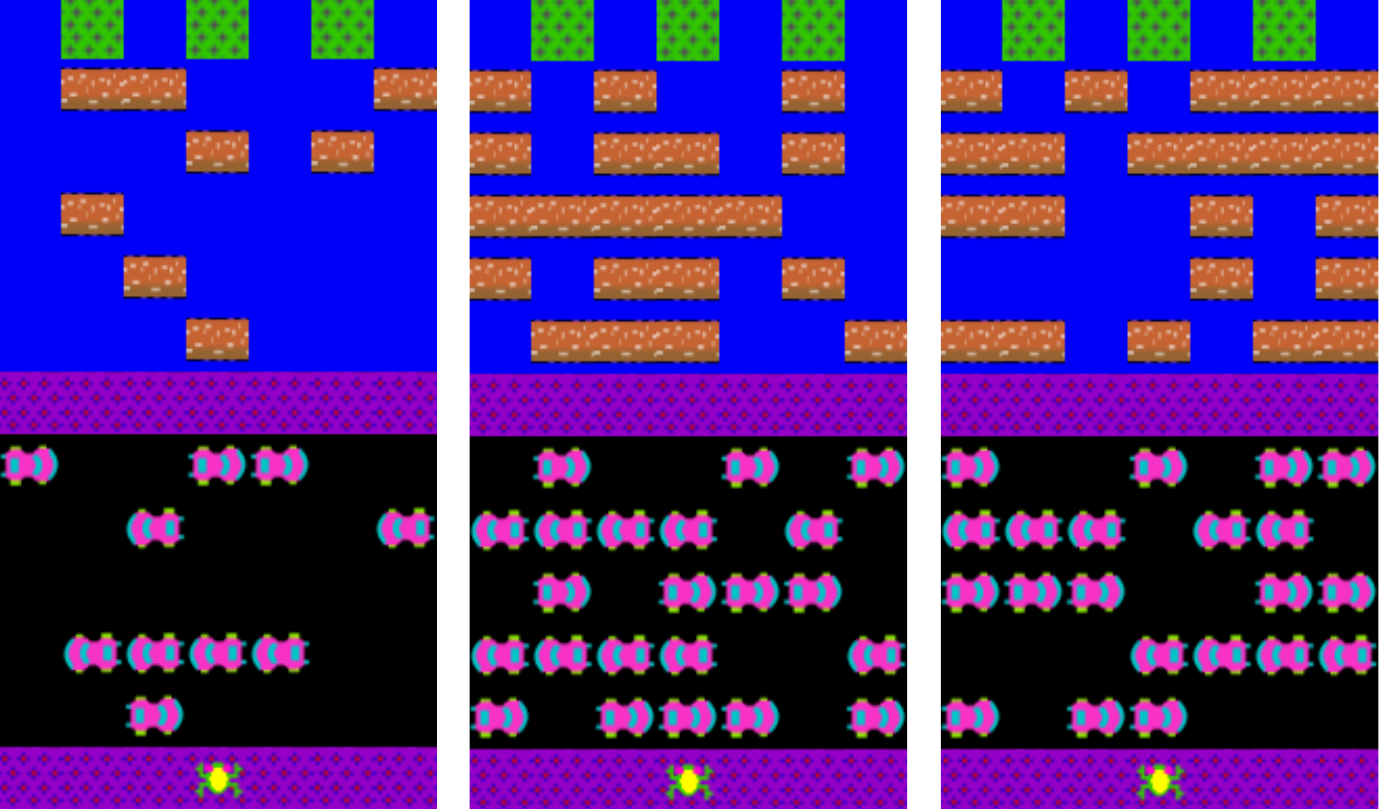}
%\captionsetup{width=.8\textwidth}
\caption{Testing and training maps made with 25\% obstacles (left), 50\% obstacles (center), and 75\% obstacles (right).}
\label{fig:maps}
\end{figure}

We conducted the experiments by generating rationalizations for states and actions in a custom implementation of the game \textit{Frogger}. 
In this environment, the agent must navigate from the bottom of the map to the top while avoiding obstacles in the environment.
The actions available to the agent in this environment are movement actions in the four cardinal directions and action for standing still.

We evaluate our rationalization technique against two baselines.
The first baseline, the \textit{random} baseline, randomly selects any sentence that can be generated by the testing grammar as a rationalization.
The second baseline, the \textit{majority vote} baseline, always selects sentences associated with the rule that is most commonly used to generate rationalizations on a given map.

Below we will discuss the process for creating the grammar, our training/test sets, and the results of this evaluation in more detail.

%\subsection{Methodology}
%\label{sec:methodology}

%\subsubsection{Environment}
%%% Mark: I don't like going this many levels deep.

% For our experiments, we chose to examine how this technique performed in the grid-based game, Frogger (Figure \ref{fig:frogger}). 
% In this environment, the agent is a frog that must navigate from the bottom of the screen to the top. 
% Scattered throughout the environment are obstacles that the agent must avoid as well as platforms that the agent must jump on in order to successfully reach its goal. 
% The actions available to the agent in this environment are movement actions in the four cardinal directions and action for standing still. 
%In this environment, the agent receives a reward of $100.0$ for reaching the other side of the map and receives a penalty of $-1.0$ for every step taken. 

\subsubsection{Grammar Creation}

In order to translate between state information and natural language, we first need ground truth rationalizations that can be associated explicitly with state and action information. 
To generate this information, we used crowdsourcing to gather a set of gameplay videos of $12$ human participants from $3$ continents playing {\em Frogger} while engaging in a think-aloud protocol, following the work by~\cite{dorst2001creativity,fonteyn1993description}. 

After players completed the game, they uploaded their gameplay video to an online speech transcription service and assigned their own utterances to specific actions. This layer of self-validation in the data collection process facilitates the robustness of the data. This process produced 225 action-rationalization trace pairs of gameplay.

%\begin{figure}
%\centering
%\begin{BVerbatim}
%"waitingAMC":["couldn't go through... #actionsWaitAMC# #waitingAMCreason#"],
%"actionsWaitAMC":["need to wait patiently for", "expecting"],
%"waitingAMCreason":["the #objects# to clear up", "#opportunity.a#"]...
%\end{BVerbatim}
%\caption{Grammar snippet of the agent's rationalization of {\em waiting} 
%\label{fig: grammar}}
%\end{figure}

We then used these action-rationalization annotations to construct a grammar for generating synthetic sentences, grounded in natural language. 
This grammar uses a set of rules based on in-game behavior of the Frogger agent to generate rationalizations that resemble the crowdsourced data gathered previously. 
% One benefit of using a grammar is that it allows us to explicitly evaluate how accurate our system is at producing appropriate rationalizations.
Since the grammar contains the rules that govern when certain rationalizations are generated, it allows us to compare automatically generated rationalizations against a ground-truth that one would not normally have if the entire training corpus was crowdsourced.

%To create the grammar, we used qualitative coding schemes and clustered the user annotation data that we gathered previously. Our technique was inspired by {\em Grounded Theory}~\cite{strauss1994grounded}, which provides an analytical tool and facilitates discovery of emerging patterns in data~\cite{walsh2015grounded}. 
%An example of the grammar can be seen in Figure \ref{fig: grammar}.
%This grammar was constructed using Tracery~\cite{compton2014tracery}. The quoted strings in this grammar represent rewrite rules that can be used to expand sentence fragments inside the brackets. In the case that there are multiple ways to expand sentences, the grammar chooses between them uniformly. %\todo{Mention and cite Tracery? Some explanation of the figure would be good: quoted strings in brackets are rewrite rules. More than rewrite rule can be given, in which case the grammar can choose uniformly.}

\subsubsection{Training and Test Set Generation}
Since we use a grammar to produce ground truth rationalizations, one can interpret the role of the encoder-decoder network as learning to reproduce the grammar. 
In order to train the network to do this, we use the grammar to generate rationalizations for each state in the environment. 
The rules that the grammar uses to generate rationalizations are based on a combination of the world state and the action taken. 
Specifically, the grammar uses the following triple to determine which rationalizations to generate: $(s_1, a, s_2)$. 
Here, $s_1$ is the initial state, $a$ is the action performed in $s_1$, and $s_2$ is the resulting state of the world after action $a$ is executed. 
States $s_1$ and $s_2$ consist of the $(x,y)$ coordinates of the agent and the current layout of grid environment. 
We use the grammar to generate a rationalization for each possible $(s_1, a, s_2)$ triple in the environment and then group these examples according to their associated grammar rules. 
For evaluation, we take $20\%$ of the examples in each of these clusters and set them aside for testing. 
This ensures that the testing set contains a representative sample of the parent population while still containing example triples associated with each rule in the grammar. 

To aid in training we duplicate the remaining training examples until the training set contains $1000$ examples per grammar rule and then inject noise into these training samples in order to help avoid overfitting.
Recall that the input to the encoder-decoder network is a triple of the form $(s_1, a, s_2)$ where $s_1$ and $s_2$ are states. 
To inject noise, we randomly select $30\%$ of the rows in this map representation for both $s_1$ and $s_2$ and redact them by replacing them with a dummy value.
%This set of examples becomes our training set.

To evaluate how our technique for rationalization performs under different environmental conditions, we developed three different maps.
The first map was randomly generated by filling 25\% of the bottom with car obstacles and filling 25\% of the top with log platforms. 
The second map was 50\% cars/logs and the third map was 75\% cars/logs (see Figure~\ref{fig:maps}). 
%See Figure \ref{fig:maps}. 
For the remainder of the paper, we refer to these maps as the \textit{25\% map}, the \textit{50\% map}, and the \textit{75\% map} respectively. 
We also ensured that it was possible to complete each of these maps to act as a loose control on map quality.

\subsubsection{Training and Testing the Network}

The parallel corpus of state-action representations and natural language are used to train an encoder-decoder neural translation algorithm based on \cite{luong-pham-manning:2015:EMNLP}. 
We use a 2-layered encoder-decoder network with attention using long short-term memory (LSTM) nodes with a hidden node size of $300$. 
We train the network for 50 epochs and then use it to generate rationalizations for each triple in the testing set.
 
To evaluate the accuracy of the encoder-decoder network, we need to have a way to associate the sentence generated by our model with a rule that exists in our grammar. 
The generative nature of encoder-decoder networks makes this difficult as its output may accurately describe the world state, but not completely align with the test example's output.
To determine the rule most likely to be associated with the generated output, we use BLEU score~\cite{papineni2002bleu} to calculate sentence similarity between the sentence generated by our predictive model with each sentence that can be generated by the grammar and record the sentence that achieves the highest score. 
We then identify which rule in the grammar could generate this sentence and use that to calculate accuracy. 
If this rule matches the rule that was used to produce the test sentence then we say that it was a match. 

Accuracy is defined as the percentage of the predictions that matched their associated test example.
We discard any predicted sentence with a BLEU score below $0.7$ when compared to the set of all generated sentences. 
% If the predicted sentence does not achieve a BLEU score of greater than $0.7$ when compared to any sentence that could be produced by the grammar, then it is automatically counted as a mismatch. 
This threshold is put in place to ensure that low quality rationalizations in terms of language syntax do not get erroneously matched to rules in the grammar. 

It is possible for a generated sentence to be associated with more than one rule in the grammar if, for example, multiple rules achieve the same, highest BLEU score. 
If the rule that generated the testing sentence matches at least one of the rules associated with the generated sentence, then we count this as a match. 
\subsubsection{Rationalization Generation Results} 
%A screenshot of an agent trained to play {\em Frogger} and ``thinking out loud'' after every action using our system is shown in Figure~\ref{fig:rationalization}.
%
The results of our experiments validating our first hypothesis can be found in Table~\ref{tab:results}. 
As can be seen in the table, the encoder-decoder network was able to consistently outperform both the random baseline and majority baseline models.
Comparing the maps to each other, the encoder-decoder network produced the highest accuracy when generating rationalizations for the 75\% map, followed by the 25\% map and the 50\% map respectively. 
To evaluate the significance of the observed differences between these models, we ran a chi-squared test between the models produced by the encoder-decoder network and random predictor as well as between the encoder-decoder network models and the majority classifier. 
Each difference was deemed to be statistically significant ($p < 0.05$) across all three maps.

\begin{table}[tb]

\begin{center}
\caption{Accuracy values for Frogger environments with different obstacle densities. Accuracy values for sentences produced by the encoder-decoder network (full) significantly outperform those generated by a random model and a majority classifier as determined by a chi-square test.}
%{\footnotesize
\begin{tabular}{ c | c  c  c}
  {\bf Map} & {\bf Full} & {\bf Random} & {\bf Majority vote}\\
  \hline
  25\% obstacles & {\bf 0.777} & 0.00 & 0.168\\
  50\% obstacles & {\bf 0.687} & 0.00 & 0.204\\
  75\% obstacles & {\bf 0.80} & 0.00 & 0.178\\
  \hline
  \end{tabular}
%  }

\end{center}
 \label{tab:results}
\end{table}

%\begin{table}[tb]
%\caption{Accuracy values for Frogger environments with different obstacle densities. Models were tested on the test sets created from different Frogger maps.}
%\begin{center}
%{\footnotesize
%\begin{tabular}{ c | c  c  c}
%  & \multicolumn{3}{c}{\bf Training states}\\
%  {\bf Testing Map} & {25\% obst.} & {50\% obst.} & {75\% obst.}\\
%  \hline
%  25\% obstacles & -- & 0.124 & 0.159\\
%  50\% obstacles & 0.453 & -- & 0.674\\
%  75\% obstacles & 0.442 & 0.732 & --\\
%  \hline
%  \end{tabular}
%  }
%%\end{center}
% \label{tab:generalization}
%\end{table}

%%%%%%%%%%%%

\subsubsection{Rationalization Generation Discussion}
The models produced by the encoder-decoder network significantly outperformed the baseline models in terms of accuracy percentage. 
This means that this network was able to better learn when it was appropriate to generate certain rationalizations when compared to the random and majority baseline models. 
Given the nature of our test set as well, this gives evidence to the claim that these models can generalize to unseen states as well.
%It is important to note, however, that claims about the possibility of these maps generalizing using the results in Table~\ref{tab:results} only extend to states drawn from maps with similar obstacle densities. 
While it is not surprising that encoder-decoder networks were able to outperform these baselines, the margin of difference between these models is worth noting. 
The performances of both the random and majority classifiers are a testament to the complexity of this problem.  

These results give strong support to our claim that our technique for creating AI rationalizations using neural machine translation can accurately produce rationalizations that are appropriate to a given situation. 

%\subsection{Understanding Human Satisfaction}
\subsection{Rationalization Satisfaction Study Methodology}
The results of our previous study indicate that our technique is effective at producing appropriate rationalizations. 
This evaluation is meant to validate our second hypothesis that humans would find rationalizations more satisfying than other types of explanation for sequential decision making problems. 
To do this, we asked people to rank and justify their relative satisfaction with explanations generated by three agents (described below) as each performs the same task in identical ways, only differing in the way they express themselves. The three agents are: 
\begin{itemize}
\item \textit{The rationalizing robot}, uses our neural translation approach to generate explanations. 
\item \textit{The action-declaring robot}, states its action without any justification. For instance, it states \enquote{I will move right}. 
\item \textit{The numerical robot}, simply outputs utility values with no natural language rationalizations. 
\end{itemize}
We will discuss our human subjects protocol and experimental results below.
% The results of our previous study indicate that our technique is effective at producing appropriate rationalizations. 
% Our second evaluation is meant to validate our second claim that humans would find rationalizations more satisfying than other types of explanation for sequential decision making problems. 
% To do this, we asked people to rank and justify their relative satisfaction with explanations generated by three agents as they each perform the same task. 
% The only difference between the agents is the way they express themselves. 
% We will discuss our methods for creating the replays as well as our human subjects protocol and experimental results below. 
% \subsubsection{Environment for Evaluation}
% To understand the dimensions of human satisfaction, we instructed people to watch replays of agents acting in Frogger. 
% We used a different Frogger environment than we used in the previous experiment (\todo {see Figure~XXX}). 
% This map was constructed to contain a better balance of obstacles and empty space, which would make it easier for humans to observe and process what was occurring in the environment. Each agent's replay took place on the same map, and each agent displayed the same navigational behavior to reach the other side of the map. 

%Here is where you would talk about the protocol}
\subsubsection{Participants}
Fifty-three adults (age range~=~22~\textendash~64 years, M~=~34.1, SD~=~9.38) were recruited from Amazon Mechanical Turk (AMT) through a management service called TurkPrime \cite{litman2017turkprime}. %\todo{cite turkprime}.  
%Participation was only offered to people between the ages of 18 and 70. %\todo{remove table and describe with words}
Twenty-one percent of the participants were women, and only three countries were reported when the participants were asked what country they reside in.
Of these, 91\% of people reported that they live in the United States. 

%Fifty-three adults (age range~=~22~\textendash~64 years, M~=~34.1, SD~=~9.38) were recruited from the Amazon Mechanical Turk (AMT) crowdsourcing web service \todo{cite turkprime}.  Participation in our study was only offered to people between the ages of 18 and 70 who had completed at least 100 tasks on AMT (and who had at least 90\% of those tasks approved). \todo{remove table and describe with words} Twenty-one percent of the participants were women.  Table~\ref{tab:country} shows that only three countries were reported when the participants were asked what country they reside in, and that most reported living in the United States.

% \begin{table}[tb]
% \caption{Distribution of current residence reported by participants}
% \begin{center}
% %{\footnotesize
% \begin{tabular}{ l | r }
%   {\bf Country} & { Count }\\
%   \hline
%   United States & { 48}\\
%   India & { 4}\\
%   Dominican Republic & { 1}\\
%   \hline
%   \end{tabular}
% %  }
% \end{center}
%  \label{tab:country}
% \end{table}

\subsubsection{Procedure}
\begin{figure}
\centering
\includegraphics[width= 3.5in]{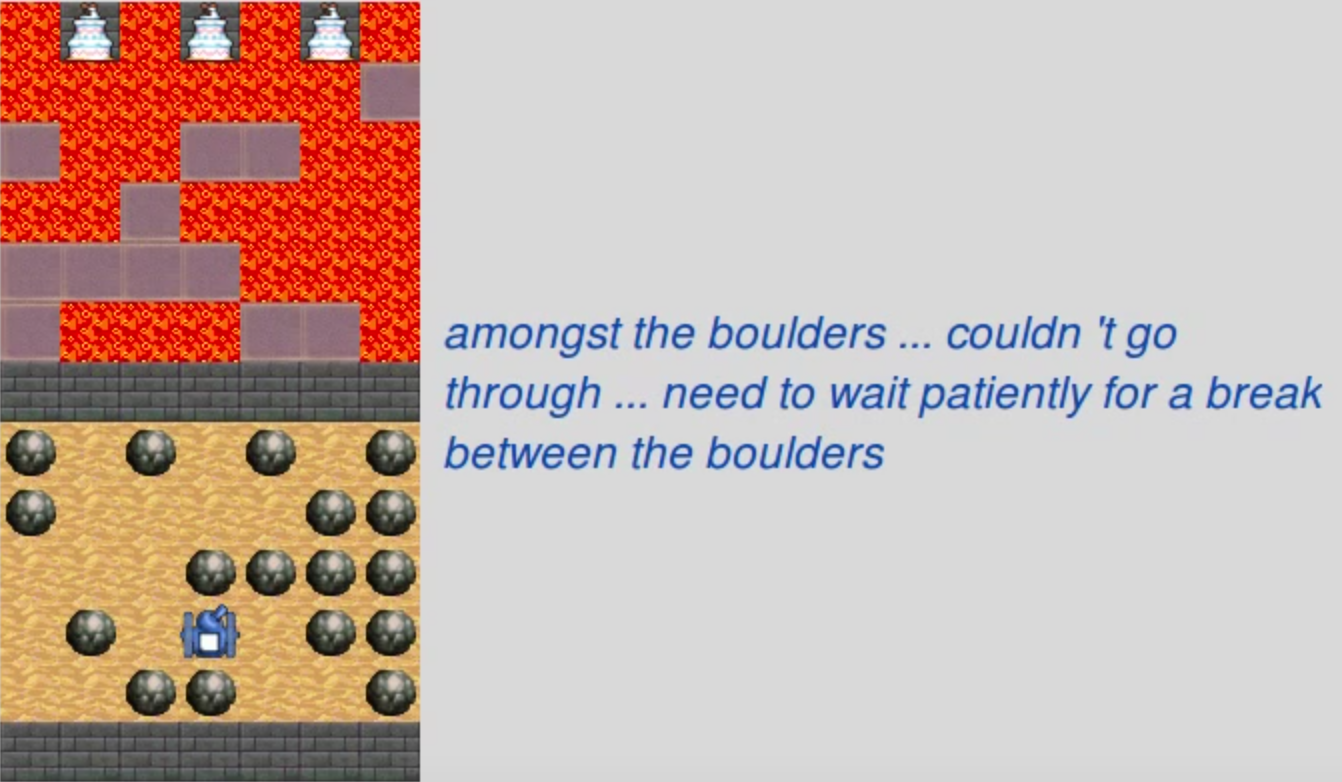}
% \includegraphics[width= 2.5in]{images/robotAexplaining_edited}
%\captionsetup{width=.8\textwidth}
\caption{The \textit{rationalizing robot} navigating the modified Frogger environment}
\label{fig:rationalize}
\end{figure}

After reading a brief description of our study and consenting to participate, participants were introduced to a hypothetical high-stakes scenario. In this scenario, the participant must remain inside a protective dome and rely on autonomous agents to retrieve food packages necessary for survival. 
The environment is essentially a \enquote{re-skinned} version of Frogger (see figure~\ref{fig:rationalize}) that is contextually appropriate for the high-stakes hypothetical scenario. To avoid effects of preconceived notions, we did not use the agents' descriptive names in the study; we introduced the agents as \enquote{Robot A} for the \textit{rationalizing} robot, \enquote{Robot B} for the \textit{action-declaring} robot, and \enquote{Robot C} for the \textit{numerical} robot.

%Three agents, referred to participants as \enquote{Robot A}, \enquote{Robot B}, and \enquote{Robot C} (see Figure~\ref{fig:lineup}), 
%\begin{figure}
%\centering
%\includegraphics[width= 3in]{images/lineup}
%\captionsetup{width=.8\textwidth}
%\caption{\todo{Added this missing figure. Please review this caption} From left to right: Robot A, Robot B, Robot C}
%\label{fig:lineup}
%\end{figure}
% Three agents navigated the same environment in identical ways, only differing in the type of explanation produced:
% \begin{itemize}
% \item \textit{The rationalizing robot}, uses our neural translation approach to generate explanations. 
% \item \textit{The action-declaring robot}, states its action without any justification. For instance, it states \enquote{I will move right}, \enquote{I will move left}, etc. 
% \item \textit{The numerical robot}, simply outputs utility values with no natural language rationalizations. 
% \end{itemize}

Next, the participants watched a series of six videos in two groups of three: three depicting the agents succeeding and three showing them failing.
Participants were quasi-randomly assigned to one of the 12 possible presentation orderings, such that each ordering was designed to have the same number of participants.
After watching the videos, participants were asked to rank their satisfaction with the expressions given by each of the three agents and to justify their choices in their own words. 

\subsubsection{Satisfaction Results and Analysis}
% \subsection{Results of Human Evaluation}

\begin{figure}
\centering
\includegraphics[width=3.5in]{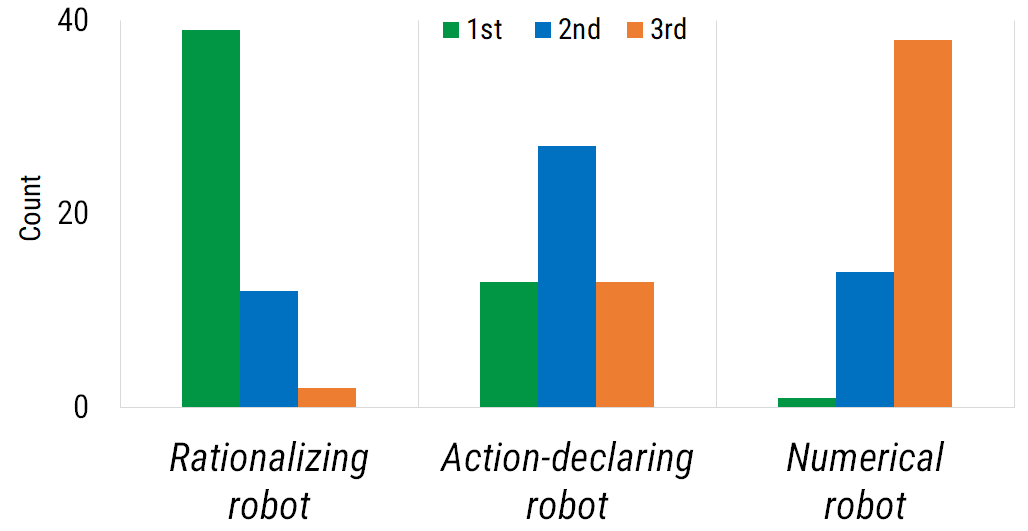}
%\captionsetup{width=.8\textwidth}
\caption{Count of $1\textsuperscript{st}$, $2\textsuperscript{nd}$, and $3\textsuperscript{rd}$ place ratings given to each robot. The rationalization robot received the most $1\textsuperscript{st}$ place, the action-declaring robot received the most $2\textsuperscript{nd}$ place, and the numeric robot received the most $3\textsuperscript{rd}$ place ratings.}
\label{fig:bars}
\end{figure}

%As a group, participants were not equally satisfied with the explanations given by the three agents.  
Figure~\ref{fig:bars} shows that the \textit{rationalizing robot} (Robot A) received the most $1\textsuperscript{st}$ place ratings, the \textit{action-declaring robot} (Robot B) received the most $2\textsuperscript{nd}$ place ratings, and the \textit{numerical robot} (Robot C) received the most $3\textsuperscript{rd}$ place ratings. To determine whether any of these differences in satisfaction ratings were significant, we conducted a non-parametric Friedman test of differences among repeated measures.  This yielded a Chi-square value of 45.481, which was significant ($p<0.001$).

To determine which of the ratings differences were significant, we made pairwise comparisons between the agents, using the Wilcoxon-Nemenyi-McDonald-Thompson test~\cite{hollander2013nonparametric}. 
%The results of our these comparisons can be found in Table~\ref{tab:satisfactionResults}. %\todo{what is the actual meaning of the pairwise comparison? instead of reporting, can we explain what it means in clear terms?}
All three comparisons yielded a significant difference in ratings. 
The satisfaction ratings for the \textit{rationalization} robot were significantly higher than those for both the \textit{action-declaring} robot ($p=0.0059$) as well as the \textit{numerical} robot ($p<0.001$). 
Furthermore, the ratings for the \textit{action-declaring} robot were significantly higher than those for the \textit{numeric} robot ($p<0.001$).

We also analyzed the justifications that participants provided for their rankings using approaches inspired by thematic analysis \cite{aronson1995pragmatic} and grounded theory \cite{strauss1994grounded}. 
Starting with an open coding scheme, we developed a set of codes that covered various reasonings behind the ranking of the robots. 
Using the codes as analytic lenses, we clustered them under emergent themes, which shed light into the dimensions of satisfaction. Through an iterative process performed until consensus was reached, we distilled the most relevant themes into insights that can be used to understand the \enquote{whys} behind satisfaction of explanations. 
In our discussion of these responses, we refer to participants using the following abbreviation: P1 is used to refer to participant 1, P2 is used to refer to participant 2, etc. 
%Participants were enumerated and referred to as P1 for Participant 1. 

%\begin{table}[tb]
%\caption{P-values for pairwise comparisons of ratings between agents.\todo{Moved suggestion to regular text and un-bolded p-value column}}
%\begin{center}
%{\footnotesize
%\begin{tabular}{ c | c }
%  {\bf Paired Comparison} & {\bf p-value }\\
%  \hline
%  Robot A vs. Robot B & {0.0059}\\
%  Robot A vs. Robot C & {$<$0.001}\\ %0.000000000023
%  Robot B vs. Robot C & {$<$0.001}\\ %0.00072
%  \hline
%  \end{tabular}
%  }
%\end{center}
% \label{tab:satisfactionResults}
%\end{table}

\subsubsection{Findings and Discussion}
%We tested whether people are differentially satisfied by agents who explain their own actions differently.  %Robot B's explanations were rated higher in satisfaction than were Robot C's, implying that natural language enhances satisfaction.
As we hypothesized, the \textit{rationalizing} agent's explanations were rated higher than were those of the other two agents, implying that rationalization enhances satisfaction over action-declaring, natural language description and over numeric expressions.
% , even though these two agents provide a more accurate description of their decision making processes.
%The open-ended responses furthermore suggest particular attributes of explanations that drive satisfaction.

%In addition to the preference for a natural language substrate, a prototypical satisfactory rationalization has 4 attributes distinguishing preference for robot A compared to B: \textit{explanatory power}, \textit{relatability},\textit{ ludic nature}, and \textit{adequately detailed}. 

%\todo{Added \enquote{emerged from thematic analysis}, as suggested. Also changed \enquote{adequately detailed} to \enquote{adequate detail}}
In addition to the preference for a natural language substrate, four attributes emerged from our thematic analysis that characterize prototypical satisfactory rationalization: \textit{explanatory power}, \textit{relatability}, \textit{ludic nature}, and \textit{adequate detail}.  
These same attributes can be used to distinguish the rationalizing robot from the action-declaring robot.

In terms of \textit{explanatory power}, the \textit{rationalizing robot's} ability to explain its actions was the most cited reasons for its superior placement in the satisfaction rankings. 
Human rationalizations allow us to form a theory of mind for the other \cite{goldman2012theory}, enabling us to better understand motivations and actions of others. 
Similarly, the rationalizing robot's ability to show participants \enquote{\ldots what it's doing and why} (P6) enabled them to \enquote{\ldots get into [the rationalizing robot's] mind} (P17), boosting satisfaction and confidence. 
Despite using natural language, the \textit{action declaring robot} yielded dissatisfaction. 
As P38 puts it, \enquote{[The action-declaring robot] explained almost nothing\ldots which was disappointing.} 
The explanatory attribute of the rationalizing robot reduces friction of communication and results in improved satisfaction. 

With respect to \textit{relatability}, the personality expressed through \textit{rationalizing robot's} explanation allowed participants to relate to it:
% Robot A exceeded the other robots in \textit{relatability} and was regarded by some participants in terms of close interpersonal relationships:
%With respect to \textit{relatability} and \textit{ludic nature}, the human-likeness of the \textit{rationalizing robot's} (A) explanation allowed participants to relate to it, improving rapport and engagement. Many were drawn to A's personality—one person commented:
\begin{quote}
[The rationalizing robot] was relatable. He felt like a friend rather than a robot. I had a connection with [it] that would not be possible with the other 2 robots because of his built-in personality. (P21)
\end{quote}
%The use of the anthropomorphic pronoun, he, instead of 'it' while referring to A can serve as a sign of kinship, which can also afford higher levels of human-AI rapport. Participants also engaged with A because of its entertaining personality, exemplified by this remark, "A was fun and entertaining. I couldn't wait to see what he would say next!" (P2)
Participants also engaged with the rationalizing robot's \textit{ludic quality}, expressing their appreciation of its perceived playfulness: \enquote{[The rationalizing robot] was fun and entertaining. I couldn't wait to see what he would say next!} (P2).

% Participants also had fun with Robot A, using \textit{ludic} terms to describe their perception of playfulness: \enquote{A was fun and entertaining. I couldn't wait to see what he would say next!} (P2) \todo{fixed quotes wherever I could find them in the document}

A rationalization yields higher satisfaction if it is \textit{adequately detailed}. 
The \textit{action-declaring robot}, despite its lack of explainability, received some positive comments. 
People who preferred the action-declaring robot over the rationalizing robot claimed that \enquote{[the rationalizing robot] talks too much} (P47), the action-declaring robot is \enquote{nice and simple} (P48), and that they \enquote{would like to experience a combination of [the action-declaring robot] and [the rationalizing robot]} (P41). Context permitting, there is a need to balance level of detail with information overload. 

These findings also align with our proposed benefits of AI Rationalization, especially in terms accessible explanations that are intuitive to the non-expert. 
We also observed how the human-centered communication style facilitates higher degrees of rapport.
The insights not only help evaluate the quality of responses generated by our system, but also sheds light into design considerations that can be used to build the next generation of explainable agents.

\section{Future Work}
\label{sec:future-work}

Our next step is to build on our current work and investigate hypotheses about how types of rationalizations impact human preferences of AI agents in terms of confidence, perceived intelligence, tolerance to failure, etc.
% \todo{@larry: pls insert that sentence you mentioned about how people's familiarity with AI can change perceptions of B over C; will most likely appear in the data from intro to AI}
To address these questions, it will be necessary to conduct experiments similar to the one described above. 
% in which participants observe or interact with autonomous systems to achieve a particular task using questionnaires to assess implications of trust and rapport.
% Questionnaires can be devised to capture and analyze participants' feelings of trust and rapport. 
% One can look at the effects of ``thinking out loud'' during execution versus rationalization when something unexpected occurs.
% Further, one can devise interventions to ensure that a failure happens or that the agent is trained to optimize a reward function that does not match operator expectations.
%Since rationalizations do not truly reflect the autonomous systems' underlying decision-making process, the question arises of how 
It will be interesting to see how inaccurate rationalizations can be before feelings of confidence and rapport are significantly affected.
Our experimental methodology can be adapted to inject increasingly more error into the rationalizations and understand human preferences.

%%%%%%%%%%%%%%%%%%%%%%%%%%%%%%%%%%%%%%%%%%%%%%%%%%%%%%%%%%%%%%%%%%%%%

\section{Conclusions}
\label{sec:conclusions}

AI rationalization provides a new lens through which we can explore the realms of Explainable AI.
% the question of what makes a good explanation of the decisions of an autonomous or semi-autonomous agent.
% As AI and society integrates further, we envision the increase in human operators (who may be non-experts) of autonomous or semi-autonomous agents. 
% who do not have computer science or AI backgrounds.
% These operators will naturally want to know why an agent fails to achieve an objective or why its behavior deviates from expectations. 
% Further, operators may need to quickly and effortlessly model the reasoning behind agent decisions in real-time scenarios.
As society and AI integrates further, we envision the increase in human operators who will want to know \textit{why} an agent does \textit{what} it does in an intuitive and accessible manner. 

% Since AI rationalization is the process of generating explanations {\em as if a human had done the behavior}, we hypothesize that rationalization will increase human operators' feelings of confidence using autonomous and semi-autonomous systems.

We have shown that creating rationalizations using neural machine translation techniques produces rationalizations with accuracies above baselines. 
We have also shown that rationalizations produced using this technique were more satisfying than other alternative means of explanation.
%Testing hypotheses about trust and rapport are discussed as next steps in determining the applicability of rationalization as an approach to explainable AI.

Rationalization allows autonomous systems to be relatable and human-like in their decision-making when their internal processes can be non-intuitive.
% decision-making processes can be very complicated and different than human reasoning.
We believe that AI rationalization can be an important step towards the democratization of real-world commercial robotic systems in healthcare, accessibility, personal services, and military teamwork.

\bibliography{bibliography}
\bibliographystyle{aaai}

\end{document}